\documentclass[10pt,twocolumn,letterpaper]{article}

\usepackage{cvpr}              % To produce the CAMERA-READY version
\usepackage[dvipsnames]{xcolor}

\definecolor{others}{RGB}{0, 0, 0}
\definecolor{nbarrier}{RGB}{255, 120, 50}
\definecolor{nbicycle}{RGB}{255, 192, 203}
\definecolor{nbus}{RGB}{255, 255, 0}
\definecolor{ncar}{RGB}{0, 150, 245}
\definecolor{nconstruct}{RGB}{0, 255, 255}
\definecolor{nmotor}{RGB}{200, 180, 0}
\definecolor{npedestrian}{RGB}{255, 0, 255}
\definecolor{ntraffic}{RGB}{255, 240, 150}
\definecolor{ntrailer}{RGB}{135, 60, 0}
\definecolor{ntruck}{RGB}{255, 0, 0}
\definecolor{ndriveable}{RGB}{213, 213, 213}
\definecolor{nother}{RGB}{139, 137, 137}
\definecolor{nsidewalk}{RGB}{75, 0, 75}
\definecolor{nterrain}{RGB}{150, 240, 80}
\definecolor{nmanmade}{RGB}{160, 32, 240}
\definecolor{nvegetation}{RGB}{0, 175, 0}

\definecolor{car}{rgb}{0.39215686, 0.58823529, 0.96078431}
\definecolor{bicycle}{rgb}{0.39215686, 0.90196078, 0.96078431}
\definecolor{motorcycle}{rgb}{0.11764706, 0.23529412, 0.58823529}
\definecolor{truck}{rgb}{0.31372549, 0.11764706, 0.70588235}
\definecolor{other-vehicle}{rgb}{0.39215686, 0.31372549, 0.98039216}
\definecolor{person}{rgb}{1.        , 0.11764706, 0.11764706}
\definecolor{bicyclist}{rgb}{1.        , 0.15686275, 0.78431373}
\definecolor{motorcyclist}{rgb}{0.58823529, 0.11764706, 0.35294118}
\definecolor{road}{rgb}{1.        , 0.        , 1.        }
\definecolor{parking}{rgb}{1.        , 0.58823529, 1.        }
\definecolor{sidewalk}{rgb}{0.29411765, 0.        , 0.29411765}
\definecolor{other-ground}{rgb}{0.68627451, 0.        , 0.29411765}
\definecolor{building}{rgb}{1.        , 0.78431373, 0.        }
\definecolor{fence}{rgb}{1.        , 0.47058824, 0.19607843}
\definecolor{vegetation}{rgb}{0.        , 0.68627451, 0.        }
\definecolor{trunk}{rgb}{0.52941176, 0.23529412, 0.        }
\definecolor{terrain}{rgb}{0.58823529, 0.94117647, 0.31372549}
\definecolor{pole}{rgb}{1.        , 0.94117647, 0.58823529}
\definecolor{traffic-sign}{rgb}{1.        , 0.        , 0.    }   

\definecolor{s_car}{rgb}{0.39215686, 0.58823529, 0.96078431}
\definecolor{s_bicycle}{rgb}{0.39215686, 0.90196078, 0.96078431}
\definecolor{s_motorcycle}{rgb}{0.11764706, 0.23529412, 0.58823529}
\definecolor{s_truck}{rgb}{0.31372549, 0.11764706, 0.70588235}
\definecolor{s_othervehicle}{rgb}{0.39215686, 0.31372549, 0.98039216}
\definecolor{s_person}{rgb}{1.        , 0.11764706, 0.11764706}
\definecolor{s_bicyclist}{rgb}{1.        , 0.15686275, 0.78431373}
\definecolor{s_motorcyclist}{rgb}{0.58823529, 0.11764706, 0.35294118}
\definecolor{s_road}{rgb}{1.        , 0.        , 1.        }
\definecolor{s_parking}{rgb}{1.        , 0.58823529, 1.        }
\definecolor{s_sidewalk}{rgb}{0.29411765, 0.        , 0.29411765}
\definecolor{s_otherground}{rgb}{0.68627451, 0.        , 0.29411765}
\definecolor{s_building}{rgb}{1.        , 0.78431373, 0.        }
\definecolor{s_fence}{rgb}{1.        , 0.47058824, 0.19607843}
\definecolor{s_vegetation}{rgb}{0.        , 0.68627451, 0.        }
\definecolor{s_trunk}{rgb}{0.52941176, 0.23529412, 0.        }
\definecolor{s_terrain}{rgb}{0.58823529, 0.94117647, 0.31372549}
\definecolor{s_pole}{rgb}{1.        , 0.94117647, 0.58823529}
\definecolor{s_trafficsign}{rgb}{1.        , 0.        , 0.        }
\definecolor{s_otherstructure}{rgb}{0.98039215, 0.58823529, 0.}
\definecolor{s_otherobject}{rgb}{0.19607843, 1.        , 1.        }

\makeatletter
\newcommand{\barrier@nuscenesfreq}{11.79}
\newcommand{\bicycle@nuscenesfreq}{0.18}
\newcommand{\bus@nuscenesfreq}{5.83}
\newcommand{\car@nuscenesfreq}{48.27}
\newcommand{\construction@nuscenesfreq}{1.92}
\newcommand{\motorcycle@nuscenesfreq}{0.54}
\newcommand{\pedestrian@nuscenesfreq}{2.93}
\newcommand{\trafficcone@nuscenesfreq}{0.93}
\newcommand{\trailer@nuscenesfreq}{6.22}
\newcommand{\truck@nuscenesfreq}{20.07}
\newcommand{\driveable@nuscenesfreq}{28.64}
\newcommand{\other@nuscenesfreq}{0.77}
\newcommand{\sidewalk@nuscenesfreq}{6.34}
\newcommand{\terrain@nuscenesfreq}{6.35}
\newcommand{\manmade@nuscenesfreq}{16.10}
\newcommand{\vegetation@nuscenesfreq}{11.08}
\newcommand{\nuscenesfreq}[1]{{\csname #1@nuscenesfreq\endcsname}}

\newcommand{\car@semkitfreq}{3.92}
\newcommand{\bicycle@semkitfreq}{0.03}
\newcommand{\motorcycle@semkitfreq}{0.03}
\newcommand{\truck@semkitfreq}{0.16}
\newcommand{\othervehicle@semkitfreq}{0.20}
\newcommand{\person@semkitfreq}{0.07}
\newcommand{\bicyclist@semkitfreq}{0.07}
\newcommand{\motorcyclist@semkitfreq}{0.05}
\newcommand{\road@semkitfreq}{15.30}  %
\newcommand{\parking@semkitfreq}{1.12}
\newcommand{\sidewalk@semkitfreq}{11.13}  %
\newcommand{\otherground@semkitfreq}{0.56}
\newcommand{\building@semkitfreq}{14.1}  %
\newcommand{\fence@semkitfreq}{3.90}
\newcommand{\vegetation@semkitfreq}{39.3}  %
\newcommand{\trunk@semkitfreq}{0.51}
\newcommand{\terrain@semkitfreq}{9.17} %
\newcommand{\pole@semkitfreq}{0.29}
\newcommand{\trafficsign@semkitfreq}{0.08}
\newcommand{\semkitfreq}[1]{{\csname #1@semkitfreq\endcsname}}
\definecolor{cvprblue}{rgb}{0.21,0.49,0.74}
\usepackage[pagebackref,breaklinks,colorlinks,citecolor=cvprblue]{hyperref}
\usepackage{amsmath}
\usepackage[accsupp]{axessibility}
\usepackage{lineno}
\usepackage{etoolbox}

\usepackage{wrapfig}
\usepackage{graphicx}
\usepackage{grffile}

\usepackage[page, header, toc, page]{appendix} 
\usepackage{hhline}
\usepackage{caption}
\usepackage{subcaption}
\usepackage{enumitem}
\usepackage[utf8]{inputenc} % allow utf-8 input
\usepackage[T1]{fontenc}    % use 8-bit T1 fonts
\usepackage{hyperref}       % hyperlinks
\usepackage{url}            % simple URL typesetting
\usepackage{booktabs}       % professional-quality tables
\usepackage{amsfonts}       % blackboard math symbols
\usepackage{nicefrac}       % compact symbols for 1/2, etc.
\usepackage{microtype}      % microtypography
\usepackage{xcolor}         % colors
\usepackage{multirow}
\usepackage{amssymb}
\usepackage{amsmath}
\usepackage{bbding}
\usepackage{makecell}
\usepackage{colortbl}  
\usepackage{array}  
\usepackage{arydshln}
\usepackage{subcaption,siunitx,booktabs}
\usepackage{empheq, nccmath}
\usepackage{tabularx}
\usepackage{mathtools}
\usepackage{adjustbox}
\usepackage{graphicx}
\usepackage{grffile}
\usepackage{tikz}
\usepackage{colortbl}
\usepackage{nicefrac}
\usepackage{wrapfig}
\usepackage{etoolbox}
\makeatletter
\pretocmd{\chapter}{\addtocontents{toc}{\protect\addvspace{-10\p@}}}{}{}
\pretocmd{\section}{\addtocontents{toc}{\protect\addvspace{-0.5\p@}}}{}{}
\pretocmd{\subsection}{\addtocontents{toc}{\protect\addvspace{-0.5\p@}}}{}{}
\makeatother
\usepackage{titletoc}
\usepackage{enumitem}
\usepackage[utf8]{inputenc} % allow utf-8 input
\usepackage[T1]{fontenc}    % use 8-bit T1 fonts
\usepackage{hyperref}       % hyperlinks
\usepackage{url}            % simple URL typesetting
\usepackage{booktabs}       % professional-quality tables
\usepackage{amsfonts}       % blackboard math symbols
\usepackage{nicefrac}       % compact symbols for 1/2, etc.
\usepackage{xcolor}         % colors
\usepackage{multirow}
\usepackage{amssymb}
\usepackage{amsmath}
\usepackage{graphicx}
\usepackage{bbding}
\usepackage{makecell}
\usepackage{colortbl}  
\usepackage{array}  
\usepackage{arydshln}
\usepackage{empheq, nccmath}
\usepackage{tabularx}
\usepackage{mathtools}
\usepackage{adjustbox}
\usepackage{algorithm}
\usepackage{algorithmic}
\usepackage{verbatim}

\DeclarePairedDelimiter\norm{\lVert}{\rVert}

\newcommand{\myparagraph}[1]{\vspace{2pt}\noindent{\bf #1}}

\definecolor{LightGrey}{rgb}{.9,.9,.9}
\definecolor{White}{rgb}{1.,0.,1.}
\definecolor{first}{rgb}{.8,.0,.0}
\definecolor{second}{rgb}{.0,.6,.0}
\definecolor{third}{rgb}{.0,.0,.8}

\def\paperID{12384} % *** Enter the Paper ID here
\def\confName{CVPR}
\def\confYear{2025}

\title{3D Occupancy Prediction with Low-Resolution Queries via\\Prototype-aware View Transformation}

\author{Gyeongrok Oh$^{1,*}$, Sungjune Kim$^{1,*}$, Heeju Ko$^1$, Hyung-gun Chi$^2$, Jinkyu Kim$^1$, Dongwook Lee$^3$, \\ Daehyun Ji$^3$, Sungjoon Choi$^1$, Sujin Jang$^{3,\dag}$, Sangpil Kim$^{1,\dag}$ \\\\
 $^1$Korea University\quad$^2$Purdue University\quad$^3$AI Center, DS Division, Samsung Electronics
 \\
}

\begin{document}
\maketitle

\def\thefootnote{*}\footnotetext{Equal contributions}

\def\thefootnote{\dag}\footnotetext{Corresponding author (s.steve.jang@samsung.com,~spk7@korea.ac.kr)}

\begin{abstract}

The resolution of voxel queries significantly influences the quality of view transformation in camera-based 3D occupancy prediction. However, computational constraints and the practical necessity for real-time deployment require smaller query resolutions, which inevitably leads to an information loss. Therefore, it is essential to encode and preserve rich visual details within limited query sizes while ensuring a comprehensive representation of 3D occupancy. To this end, we introduce \textit{ProtoOcc}, a novel occupancy network that leverages prototypes of clustered image segments in view transformation to enhance low-resolution context. In particular, the mapping of 2D prototypes onto 3D voxel queries encodes high-level visual geometries and complements the loss of spatial information from reduced query resolutions. Additionally, we design a multi-perspective decoding strategy to efficiently disentangle the densely compressed visual cues into a high-dimensional 3D occupancy scene. Experimental results on both Occ3D and SemanticKITTI benchmarks demonstrate the effectiveness of the proposed method, showing clear improvements over the baselines. More importantly, ProtoOcc achieves competitive performance against the baselines even with 75\% reduced voxel resolution. \textit{Project page}: \url{https://kuai-lab.github.io/cvpr2025protoocc}.
\end{abstract}

\vspace{-1em}
\section{Introduction}
\label{sec:intro}
3D occupancy prediction (\textit{3DOP}) is the task of determining which parts of a 3D space are occupied by objects and identifying their class categories. In particular, view transformation from 2D to 3D space is an essential step for successful camera-based occupancy prediction. Prior works employ various 3D space representation strategies, such as bird’s-eye-view (BEV) plane, tri-perspective-view (TPV) planes, and 3D voxel cube, which aim to map multi-view camera information into corresponding grid cells in a unified 3D space. Among these, 3D voxels are the most common representation strategy in the 3DOP task~\cite{huang2023tri, wang2023panoocc, huang2023selfocc, wei2023surroundocc, zhang2023occformer, tong2023scene, vobecky2024pop, zhang2024radocc}, as they naturally encode visual information into a structured semantic 3D space. Therefore, the effectiveness of 3D voxelization (\ie, voxel queries) plays a crucial role in determining the prediction performance in camera-based 3DOP.

\begin{figure}[t]
\begin{center}
   \includegraphics[width=\linewidth]{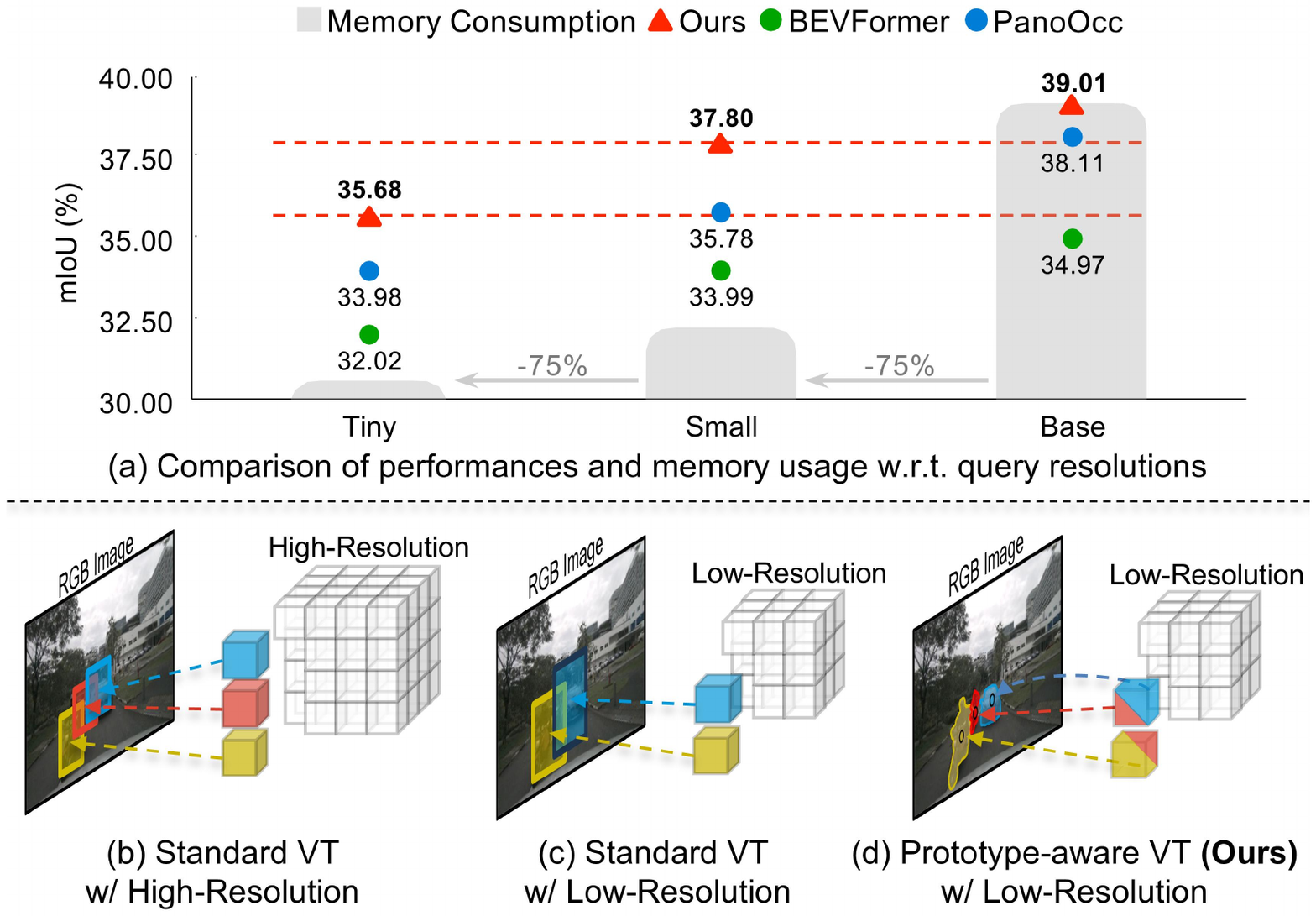}
\end{center}
\vspace{-1.0em}
\caption{(a)~Our \textit{ProtoOcc} can perform comparably to higher-resolution counterparts while using 75\% less memory. (b-c) Reducing query resolutions in standard view transformation~(VT) is required for faster inference, but brings geometrical ambiguity. (d) Our prototype-aware VT can capture high-level geometric details while preserving computational efficiency.}

\label{fig:fig_motiv}
\vspace{-1.5em}
\end{figure}
% \textbf{Motivation.} 

Trade-offs between computation and performance have long been a challenge in various deep learning applications~\cite{liu2018dynamic, kochkov2021machine, freire2021performance, hong2021lpsnet}. Learning voxel queries also encounters this dilemma, especially in real-time and safety-critical vision systems such as autonomous driving and robotics. As illustrated in Figure~\ref{fig:fig_motiv}, utilizing voxel queries with high-resolution may produce reliable performances, but requires intensive computation. Accelerating the inference inevitably necessitates smaller query sizes. However, it results in performance degradation primarily due to its inability to capture precise high-level details within limited spatial storage, which is crucial for 3DOP. Hence, incorporating comprehensive visual contexts into low-resolution voxel queries is essential for ensuring accurate 3DOP results while maintaining computational efficiency.

Despite its importance, existing approaches overlook the constraints posed by low-resolution queries in view transformation. They typically utilize a standard 2D-to-3D cross attention mechanism using low-level image features~\cite{huang2023tri, li2022bevformer, liu2022petr, wang2022detr3d, liu2023petrv2}. However, with reduced query resolutions, it becomes insufficient to solely rely on these low-level features to precisely reconstruct a 3D scene. Specifically, since images are packed into a smaller space, the encoded features lose spatial distinctiveness, thereby requiring an alternative method to preserve the necessary contextual information. In addition, it is also challenging to decode the densely compressed and low-resolution voxel queries into a sparse and high-resolution 3D occupancy scene.

To this end, we propose \textit{ProtoOcc}, a novel occupancy network focusing on context enhancement in low-resolution view transformation for 3DOP. In particular, ProtoOcc features two main learning strategies: Prototype-aware View Transformation and Multi-perspective Occupancy Decoding.

\myparagraph{Prototype-aware View Transformation.}
 Image prototypes are representations of clustered image segments that provide a high-level understanding of visual structures~\cite{wang2023learning, liang2024clusterfomer}~(e.g. layouts and boundaries). Therefore, mapping these structural elements of 2D onto 3D voxel queries effectively alleviates the loss of spatial cues in reduced query resolutions. Furthermore, we optimize the prototype features to gain distinctiveness by leveraging pseudo-mask-based contrastive learning. As a result, each prototype becomes more exclusive and informative, enabling the voxel queries to encode more precise spatial details even with reduced resolutions.

\myparagraph{Multi-perspective Occupancy Decoding.}
Reconstructing 3D occupancy from low-resolution queries is an ill-posed problem, requiring precise disentanglement of densely encoded visual details. Therefore, we propose a multi-perspective decoding strategy using voxel augmentations. Each augmentation is upsampled as a unique view of the scene, collectively forming a comprehensive representation of the 3D structure. Then, we apply a consistency regularization on different perspectives to ensure that they represent the same 3D scene, which brings robustness in training.

Through extensive experiments on the Occ3D-nuScenes~\cite{tian2024occ3d} and SemanticKITTI~\cite{behley2019semantickitti} benchmarks, we validate that the proposed ProtoOcc can be an effective solution for ensuring accurate 3DOP results while maintaining computational efficiency. Most importantly, ProtoOcc achieves competitive performance with even 75\% smaller voxel queries against models with larger-sized ones.

In summary, the contributions of this work include:
\begin{itemize}
 \item \noindent Introducing ProtoOcc as an exemplar in 3DOP by using computationally efficient low-resolution voxel queries.
 \item \noindent A prototype-aware view transformation and multi-perspective decoding strategy for enhancing the representations of low-resolution voxel queries.
 \item \noindent Demonstrating clear improvements over previous state-of-the-art methods in two major benchmarks, along with detailed analyses.
\end{itemize}

\section{Related Work}
% \label{sec:related}

% \myparagraph{Camera-based 3D Occupancy Prediction.}
\subsection{Camera-based 3D Occupancy Prediction}
 The task of camera-based 3D occupancy prediction~(3DOP) originated from Semantic Scene Completion~(SSC)~\cite{cao2022monoscene, song2017semantic, zhang2018efficient, li2020anisotropic, jiang2024symphonize}, which leverages a single monocular image or sparse LiDAR supervision of the SemanticKITTI benchmark~\cite{behley2019semantickitti} to reconstruct a static 3D scene. The monocular camera-based methods MonoScene~\cite{cao2022monoscene} and VoxFormer~\cite{li2023voxformer} serve as the foundation for further research expansion into multi-view images. Following the previous advances, Occ3D~\cite{tian2024occ3d} recently released a benchmark for dynamic 3DOP using multi-view camera images, triggering the popularity of camera-based 3DOP~\cite{huang2023tri, wang2023panoocc, huang2023selfocc, wei2023surroundocc, zhang2023occformer, tong2023scene, vobecky2024pop}.

Despite their potential, the tremendous computation load of intermediate voxel representations poses a significant challenge. Several approaches~\cite{tian2024occ3d, mei2023camera, liu2023fully} sparsify the voxel queries based on occupancy scores. A recent work COTR~\cite{ma2023cotr} first embeds 2D observations into a larger-sized voxel queries and further downsamples them into a compact representation. However, our proposed ProtoOcc directly exploits a smaller-sized query for view transformation. Our prototype-aware view transformation and multi-perspective voxel decoding strategies on low-resolution queries efficiently improve both computation and performance.

\begin{figure*}[ht]
\begin{center}
\includegraphics[width=\linewidth]{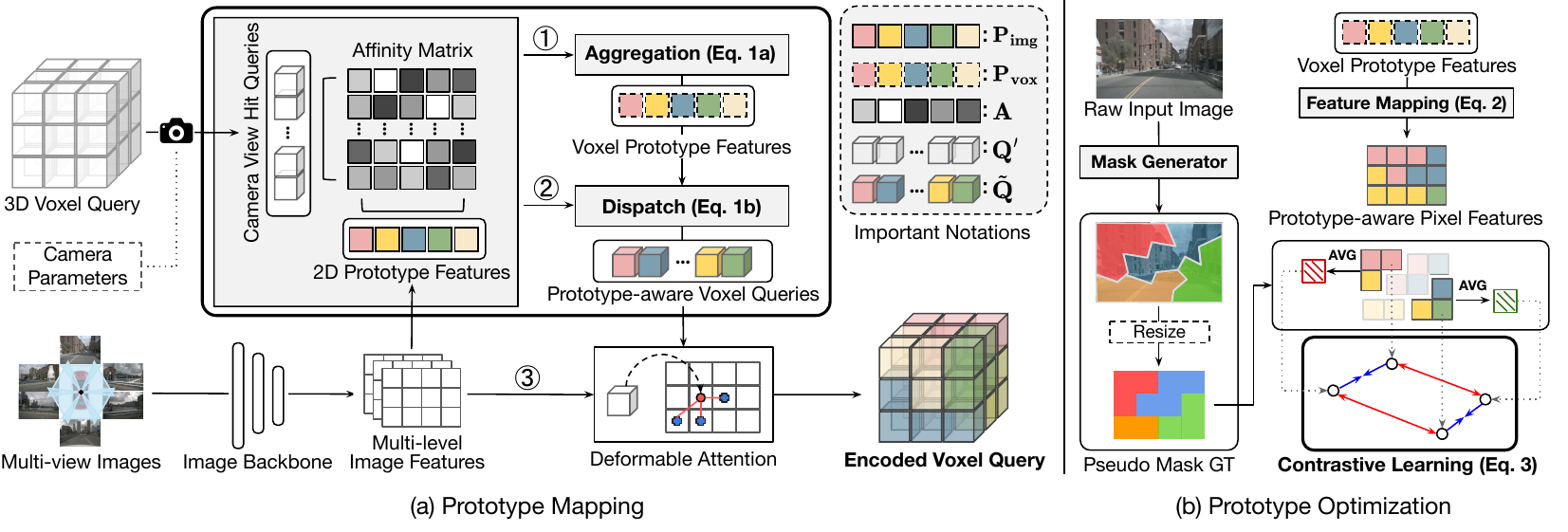}
\end{center}
\vspace{-1.0em}
\caption{\textbf{Prototype-aware View Transformation.} (a) In the Prototype Mapping stage, we fully exploit the hierarchies of 2D image features via a clustering method to map 2D prototype representations onto 3D voxel query. (b) Contrastive learning on the prototype features based on the pseudo ground truth masks enhances the discrimination between the prototypes for better feature learning. Best viewed in color.}
\label{fig:selective}
\vspace{-0.5em}
\end{figure*}

% \myparagraph{2D to 3D View Transformation.}
\subsection{2D-to-3D View Transformation}
View transformation is an indispensable process in camera-based perception models, bridging the cross-view environment. LSS~\cite{philion2020lift} achieves this by utilizing camera parameters to estimate the depth distribution from an image. The strategy is adopted from diverse 3D downstream tasks, demonstrating its effectiveness~\cite{huang2021bevdet, huang2022bevdet4d, li2023bevdepth, ma2023cotr}. However, the reliance on depth estimation causes geometric uncertainty, thus presenting a fundamental challenge. The attention-based approach is another line of view transformation that effectively aggregates 2D spatio-temporal features~\cite{li2022bevformer, kim2024enhanced,roh2022ora3d}. Here, learnable queries with diverse configurations~\cite{ li2022bevformer, liu2022petr, huang2023tri} attend on specific regions in an image and transfers 2D semantics into a unified 3D space. In 3DOP, 3D voxel queries are routinely exploited for their ability to preserve as much spatial information of a 3D scene~\cite{wang2023panoocc, tian2024occ3d}.

ProtoOcc aims at providing high-level 2D contextual guidance in attention-based view transformation, specifically targeted at enhancing the representations of low-resolution voxel queries for efficient 3DOP. Several attempts have been previously made in 3D object detection task, to improve representations in view transformation~\cite{li2023dfa3d, wang2023frustumformer}. However, these methods mainly focus on enhancing object-wise representations, which makes them unsuitable for 3DOP, which requires holistic understanding of a scene.

\section{Method}

ProtoOcc tackles the information loss problem occurring from reduced query resolutions through two main strategies: Prototype-aware View Transformation and Multi-perspective Occupancy Decoding. In this section, we first go over the overview of ProtoOcc (Sec. \ref{sec:overview}), then explain the technical details of each proposed method (Sec. \ref{sec:encode} and \ref{sec:decode}).

\subsection{Overview}
\label{sec:overview}
Given $N$ multi-view images $\{\mathcal{I}_{i}\}_{i=1}^{N}$, camera-based 3D occupancy prediction aims to reconstruct a 3D occupancy scene $\mathcal{O}\in \mathbb{R}^{L\times H\times W\times Z}$, where $H$, $W$, and $Z$ represent the spatial dimensions and $L$ is the semantic label. Specifically, the image backbone network~(e.g. ResNet50~\cite{he2016deep}) first extracts multi-scale image features $\{\{\mathcal{F}_{i}^{(t)}\}_{i=1}^{N}\}_{t=1}^{T}$ using Feature Pyramid Network~(FPN)~\cite{lin2017feature}. Here, $T$ denotes the total number of different feature scales, and each feature has a channel size of $d$. Then, these 2D features across different views are converted into a unified 3D representation by employing a learnable 3D voxel query $\mathbf{Q}\in\mathbb{R}^{d\times h\times w\times z}$. $\mathbf{Q'}\in \mathbb{R}^{N\times K\times d}$ denotes a subset of $\mathbf{Q}$, where $K$ represents the maximum number of hit queries on each $N$ camera views~($0\leq K\leq h\times w\times z$). Following the attention-based view transformation~\cite{wang2023panoocc, li2022bevformer, tian2024occ3d}, the queries aggregate spatial and temporal cue of the surrounding 3D scene via deformable attention~\cite{zhu2020deformable}. Finally, the voxel upsampling network restores high-resolution voxel volume from the encoded voxel query and predicts an occupancy state and semantic label for each voxel.

\subsection{Prototype-aware View Transformation}
\label{sec:encode}
In this stage, we encode prototype representations of 2D images into 3D voxel queries, which is an unexplored concept in the previous literature. Prototypes are grouped representations of image clusters that integrate semantically similar features~\cite{achanta2012slic, jampani2018superpixel, yang2020superpixel}. 
% Clustering algorithms integrate semantically similar features and generate grouped representations~(i.e. prototypes) within a limited space~\cite{achanta2012slic, jampani2018superpixel, yang2020superpixel}.
Inspired by this principle, we enrich the 3D voxel query representations by mapping hierarchical image representation encompassing both raw and clustered features, thereby compensating for the loss of spatial cues at reduced resolutions. We provide an illustration of our method in Figure~\ref{fig:selective}.

% \myparagraph{Prototype Mapping.}
\subsubsection{Prototype Mapping}
Mapping image prototype representations onto 3D voxel queries starts from clustering the images. We adopt an iterative grouping strategy~\cite{achanta2012slic, jampani2018superpixel, yang2020superpixel, ding2023hgformer} to segment the image feature space. Specifically, we first divide the feature map into regular grids and obtain initial prototype representation by calculating the average within each grid. Then, we iteratively update this feature by multiplying the soft-assigned pixel-prototype similarity with the feature map. The acquired 2D prototype features are denoted as $\mathbf{P}_{\text{img}}\in \mathbb{R}^{N\times M\times d}$, where $M$ is the number of prototypes.

The next step is to lift these 2D prototype features $\mathbf{P}_{\text{img}}$ into the 3D voxel query space. In this process, we introduce an innovative application of \textit{Feature Aggregation \& Dispatch} technique~\cite{liang2024clusterfomer, ma2023image}. Originally, the idea is introduced as a feature extraction paradigm in 2D image space. In this work, we utilize this as a mapping function from a 2D prototype feature space to a 3D voxel query space. We start by projecting $\mathbf{P}_{\text{img}}$ to 3D voxel feature space using MLP layers and obtain $\hat{\mathbf{P}}_{\text{img}}$. Then, we compute the affinities $\mathbf{A}\in \mathbb{R}^{N\times M\times K}$ via pairwise cosine similarity between $\hat{\mathbf{P}}_{\text{img}}$ and $\mathbf{Q'}$. 
% Here, $\mathbf{Q'}$ denotes a subset of $\mathbf{Q}$, where $K$ represents the maximum number of hit queries on each $N$ camera views~($0\leq K\leq h\times w\times z$).
Using the sigmoid function $\sigma$, the similarity values are further re-scaled to range between (0, 1), indicating the probability of each query cell in $\mathbf{Q'}$ being assigned to a certain prototype. Multiplying these probability with the voxel query feature $\mathbf{Q'}$ produces 3D voxel prototype feature $\mathbf{P}_{\text{vox}}\in \mathbb{R}^{N\times M\times d}$~(\textit{Aggregation}). Then, these features are redistributed to individual voxel queries to form a prototype-aware 3D voxel query $\tilde{\mathbf{Q}}\in \mathbb{R}^{N\times K\times d}$~(\textit{Dispatch}). The overall process is formulated as follows:
\begin{subequations}
    \begin{eqnarray}
        \texttt{Aggregate:}\; \mathbf{P}_{\text{vox}} = {1\over{R}}\Big(\hat{\mathbf{P}}_{\text{img}}+\sigma(\mathbf{A})\cdot \mathbf{Q'} \Big) 
        \label{eq:2a} \\
        \texttt{Dispatch:}\; \tilde{\mathbf{Q}}=\mathbf{Q'} + \texttt{MLP}\Big(\sigma(\mathbf{A})^{\mathsf{T}} \cdot \mathbf{P}_{\text{vox}} \Big),
        \label{eq:2b}
    \end{eqnarray}
    \label{eqangulos}
\end{subequations}
where $\hat{\mathbf{P}}_{\text{img}}$ and $\mathbf{Q'}$ are added as residual connection for \textit{Aggregation} and \textit{Dispatch}, respectively. $R$ is the total sum of the similarities $\sigma(\mathbf{A})$, which is divided to ensure a stable training process. Subsequently, the prototype-aware voxel query $\tilde{\mathbf{Q}}$ and the multi-scale image feature $\mathcal{F}$ are fed as inputs for deformable attention. Owing to this strategy, the encoded 3D voxel query carries both high-level and fine-grained visual contexts that are highly essential for predicting 3D occupancy.

% \myparagraph{Prototype Optimization.}
\subsubsection{Prototype Optimization}
The prototype-aware 3D voxel query contains rich 2D contextual information, yet the quality of these features highly depends on 2D clustering quality. However, standard optimization objectives in 3DOP~(e.g., $\mathcal{L}_{\text{occ}}$ and $\mathcal{L}_{\text{Lov}}$) do not provide suitable guidance for cluster learning. Therefore, we incorporate an explicit pseudo 2D supervision, by a contrastive learning between the prototype-aware pixel features in accordance with pseudo ground truth masks.

\myparagraph{Prototype-aware Pixel Features.}
These features, denoted as $\mathbf{X}\in \mathbb{R}^{N\times d\times h'\times w'}$, are the mapping of 3D voxel prototype features $\mathbf{P}_\text{vox}$ onto an implicit 2D grid cell. For simplicity, we flatten the spatial dimension and denote it as $D$ in the following~(i.e., $D=h'\times w'$). Since $\mathbf{P}_\text{vox}$ itself does not contain any spatial information, a direct mapping between the two spaces is limited. Therefore, we first employ a deformable attention map~\cite{roh2021sparse} $\mathbf{G}\in\mathbb{R}^{N\times D}$, which highlights the salience regions on the image features for each 3D voxel queries. Then, the element-wise multiplication between $\mathbf{G}$ and the prototype-query affinity matrix $\mathbf{A}$ computes the probability of each grid cell being assigned with a certain prototype. This acts as the spatial bridge for mapping $\mathbf{P}_\text{vox}$ onto the 2D grid cell. Consequently, through a matrix multiplication between this feature and $\mathbf{P}_\text{vox}$, we obtain the prototype-aware pixel features. The explained mapping procedure can be formulated as:
\begin{equation}
\label{eq:p}
    \mathbf{X}=\big\{\mathbf{G}\odot \mathcal{H}(\mathbf{A})\big\}* \mathbf{P}_\text{vox},
\end{equation}
where $\mathcal{H}$ is a linear mapping function and $\odot$ is the element-wise multiplication. Through this proposed methodology, each feature in $\mathbf{X}$ stores the prototype features from the 3D voxel space, yet precisely aligned within the 2D pixel grid.

\myparagraph{Contrastive Learning with Pseudo Masks.}
The absence of explicit ground truth to contrast the prototype-aware pixel features remains a challenge. Therefore, we generate $S$ pseudo ground truth masks by applying either a classical clustering algorithm~\cite{van2012seeds} or visual foundation model~\cite{kirillov2023segment} on the image. This is further resized to match the spatial dimension of $\mathbf{X}$ to provide distinct spatial boundaries. Then, we compute $S$ mask centroids by averaging the features of $\mathbf{X}$ that fall within each boundary. These mask centroids, denoted as $\{\mathbf{M}_{i}\}_{i=1}^{S}\in \mathbb{R}^{d}$, serve as the anchor for our contrastive learning. We calculate the contrastive loss~\cite{Van_Gansbeke_2021_ICCV, Wang_2021_CVPR} for the feature in the $(x,y)$ position of $\mathbf{X}$ as:
\begin{equation}
\label{eq:3}
    \mathcal{L}_{\text{cls}}^{(x,y)}=-\log{{\sum_{s=1}^{S} m_{s} \exp{(\langle \mathbf{M}_{s}, \mathbf{X}_{(x,y)}\rangle/\tau_{\text{cls}})}}\over{\sum_{s=1}^{S} \exp{(\langle \mathbf{M}_{s}, \mathbf{X}_{(x,y)}\rangle/\tau_{\text{cls}})}}},
\end{equation}
where $\langle\cdot,\cdot\rangle$ represents the cosine similarity between the two elements, and $\tau_{\text{cls}}$ is the temperature hyperparameter. The binary value $m_{s}\in \{0,1\}$ is set as 1 if the position $(x,y)$ lies within the boundary of $s^{th}$ mask. By summing up the losses across all grid cells, we obtain the final contrastive loss~(i.e. $\mathcal{L}_{\text{cls}}=\sum_{x=1}^{h'}\sum_{y=1}^{w'}\mathcal{L}_{\text{cls}}^{(x,y)}$), which enhances the distinctiveness of the prototype features when minimized.

\subsection{Multi-Perspective Occupancy Decoding} \label{sec:decode}
When reconstructing 3D occupancy from low-resolution queries, the lack of resolution causes inevitable geometrical ambiguity. That being said, an encoded query has the potential to be decoded in diverse 3D perspectives. We address this ill-posed property by enhancing contextual diversity in occupancy decoding through two essential techniques: 1) Multi-perspective View Generation and 2) Scene Consistency Regularization.

\subsubsection{Multi-perspective View Generation}
We generate multi-perspective contexts from voxel queries through augmentations.
Since augmentation on voxel representations is not widely explored, we establish two categories of voxel augmentation: 1) a feature-level augmentation~(e.g. Random Dropout and Gaussian Noise), and 2) a spatial-level augmentation~(e.g. Transpose and Flips). Combining the augmentations from these categories, we obtain a query set $\mathbb{Q}=\{\mathbf{Q}^{(0)}, \mathbf{Q}^{(1)}, \ldots ,\mathbf{Q}^{(P-1)},\mathbf{Q}^{(P)}\}$, where $\mathbf{Q}^{(0)}$ denotes the original voxel query derived from the encoder, and $P$ is the number of voxel augmentations.
Then, these queries pass through a transposed 3D convolution layer with shared weights, generating an upsampled query set $\mathbb{V}=\{\textbf{V}^{(0)}, \textbf{V}^{(1)}, \ldots ,\textbf{V}^{(P-1)},\textbf{V}^{(P)}\}$, where 
$\textbf{V}\in \mathbb{R}^{L\times H\times W\times Z}, \forall \textbf{V}\in \mathbb{V}$.
For each grid cell position, the shared kernel convolves with diverse features of local  voxel neighbors. As a result, the upsampled voxels can be interpreted as unique perspective of a
scene, collectively forming a holistic representation of the 3D structure.

\subsubsection{Scene Consistency Regularization}
Although the upsampled query set $\mathbb{V}$ contains varying contextual perspectives, they should all depict the same semantic occupancy. Therefore, we couple with the augmentations a regularization term to maintain a semantically consistent prediction.
Specifically, we adopt a simple yet effective regularization technique of GRAND~\cite{feng2020graph}, and minimize the $L_{2}$ distances between the predicted label distributions and their average distribution for each grid cell. For example, the average of the predicted label distribution in the $(i,j,k)$ position of the voxels is calculated as $\hat{\mathbf{V}}_{(i,j,k)}=\frac{1}{P+1}\sum_{p=0}^{P} \mathcal{G}(\mathbf{V}^{(p)}_{(i,j,k)})$, where $\mathcal{G}(\cdot)$ is the label classifier network. Then, this distribution is sharpened as:
\begin{equation}
\vspace{-0.3em}
    \Tilde{\mathbf{V}}_{(i,j,k)}[c]=\hat{\mathbf{V}}^{1\over\tau_{\text{cons}}}_{(i,j,k)}[c]\bigg/ \sum_{l=1}^{L} \hat{\mathbf{V}}^{1\over\tau_{\text{cons}}}_{(i,j,k)}[l], (1\leq c \leq L),
    \vspace{-0.3em}
\end{equation}
which denotes the guessed probability on the $c^{th}$ class, with $\tau_{\text{cons}}$ as the temperature hyperparmeter that controls the sharpness of the categorical distribution. Accordingly, the final consistency regularization loss $\mathcal{L}_{\text{cons}}$ is obtained  by taking the  average of the distances between the sharpened average and each prediction across all grid cells and augmentations:
\begin{equation}
\label{eq:5}
\mathcal{L}_{\text{cons}}=\frac{\sum\limits_{p=0}^{P}\sum\limits_{k=1}^{Z}\sum\limits_{j=1}^{W}\sum\limits_{i=1}^{H}\norm[\Big]{\Tilde{\mathbf{V}}_{(i,j,k)}-\mathcal{G}(\mathbf{V}^{(p)}_{(i,j,k)})}^{2}_{2}}{H\cdot W\cdot Z\cdot (P+1)} .
% \vspace{-0.5em}
\end{equation}

\subsection{Optimization}
ProtoOcc minimizes four loss functions for each objectives: 1) $\mathcal{L}_{\text{occ}}$ for accurate semantic label prediction, 2) $\mathcal{L}_{\text{Lov}}$ as an complementary loss for maximizing the mean Intersection over Union~(mIoU), 3) $\mathcal{L}_{\text{cls}}$ for enhancing the prototype feature quality, and lastly 4) $\mathcal{L}_{\text{cons}}$ for regularizing the outputs from diverse contextual perspectives. 
The ﬁnal objective function of the model is defined as:
\vspace{-0.2em}
\begin{equation}
    \mathcal{L}_{\text{total}}=\sum_{p=0}^{P}(\lambda_{1}\mathcal{L}_{\text{occ}}^{(p)}+\lambda_{2}\mathcal{L}_{\text{Lov}}^{(p)})+\lambda_{3}\mathcal{L}_{\text{cls}}+\lambda_{4}\mathcal{L}_{\text{cons}},
% \vspace{-1em}
\end{equation}
where $\lambda_{1}$, $\lambda_{2}$, $\lambda_{3}$, and $\lambda_{4}$ are coefficient hyperparameters that determine the weights of four losses. The values we set for each coefficient are specified in Appendix~B.1.

\begin{table*}[t!]
		\scriptsize
		\setlength{\tabcolsep}{0.0068\linewidth}
		
		\newcommand{\classfreq}[1]{{~\tiny(\nuscenesfreq{#1}\%)}}  %
		\centering
		\caption{\textbf{Quantitative results on the Occ3D-nuScenes~\cite{tian2024occ3d} validation set.} We highlight the best and runner-up results for each category in \textbf{bold} and plain, respectively. Not only does ProtoOcc stand out in its category, but also by comparing the results within the same color marks~($\textcolor{red}{\bullet}$ and $\textcolor{ForestGreen}{\bullet}$), it is apparent that ProtoOcc can overcome query deficiencies, performing on par even with higher-resolution counterparts.}
        \vspace{-0.2em}
            % \scalebox{0.98}{
		\begin{tabular}{c|c| c | c c c c c c c c c c c c c c c c c}
			\toprule
			\shortstack{Query \\ Size} & \makecell{Model} & \makecell{mIoU}
                & \rotatebox{90}{\textcolor{others}{$\blacksquare$} others}
			& \rotatebox{90}{\textcolor{nbarrier}{$\blacksquare$} barrier}
			& \rotatebox{90}{\textcolor{nbicycle}{$\blacksquare$} bicycle}
			& \rotatebox{90}{\textcolor{nbus}{$\blacksquare$} bus}
			& \rotatebox{90}{\textcolor{ncar}{$\blacksquare$} car}
			& \rotatebox{90}{\textcolor{nconstruct}{$\blacksquare$} const. veh.}
			& \rotatebox{90}{\textcolor{nmotor}{$\blacksquare$} motorcycle}
			& \rotatebox{90}{\textcolor{npedestrian}{$\blacksquare$} pedestrian}
			& \rotatebox{90}{\textcolor{ntraffic}{$\blacksquare$} traffic cone}
			& \rotatebox{90}{\textcolor{ntrailer}{$\blacksquare$} trailer}
			& \rotatebox{90}{\textcolor{ntruck}{$\blacksquare$} truck}
			& \rotatebox{90}{\textcolor{ndriveable}{$\blacksquare$} drive. suf.}
			& \rotatebox{90}{\textcolor{nother}{$\blacksquare$} other flat}
			& \rotatebox{90}{\textcolor{nsidewalk}{$\blacksquare$} sidewalk}
			& \rotatebox{90}{\textcolor{nterrain}{$\blacksquare$} terrain}
			& \rotatebox{90}{\textcolor{nmanmade}{$\blacksquare$} manmade}
			& \rotatebox{90}{\textcolor{nvegetation}{$\blacksquare$} vegetation}
			\\
   
			\midrule

                \multirow{7}{*}{\shortstack{\texttt{Base}}}

                & CTF-Occ~\cite{tian2024occ3d} & \textcolor{lightgray}{28.50} & \textcolor{lightgray}{8.09} & \textcolor{lightgray}{39.33} & \textcolor{lightgray}{20.56} & \textcolor{lightgray}{38.29} & \textcolor{lightgray}{42.24} & \textcolor{lightgray}{16.93} & \textcolor{lightgray}{24.52} & \textcolor{lightgray}{22.72} & \textcolor{lightgray}{21.05} & \textcolor{lightgray}{22.98} & \textcolor{lightgray}{31.11} & \textcolor{lightgray}{53.33} & \textcolor{lightgray}{33.84} & \textcolor{lightgray}{37.98} & \textcolor{lightgray}{33.23} & \textcolor{lightgray}{20.79} & \textcolor{lightgray}{18.00} \\ %

                & TPVFormer~\cite{huang2023tri} &  \textcolor{lightgray}{34.20} & \textcolor{lightgray}{7.68} & \textcolor{lightgray}{44.01} & \textcolor{lightgray}{17.66} & \textcolor{lightgray}{40.88} & \textcolor{lightgray}{46.98} & \textcolor{lightgray}{15.06} & \textcolor{lightgray}{20.54} & 24.69 & 24.66 & \textcolor{lightgray}{24.26} & \textcolor{lightgray}{29.28} & \textcolor{lightgray}{79.27} & \textcolor{lightgray}{40.65} & \textcolor{lightgray}{48.49} & \textcolor{lightgray}{49.44} & \textcolor{lightgray}{32.63} & \textcolor{lightgray}{29.82}   \\ %

                & SurroundOcc~\cite{wei2023surroundocc} &  \textcolor{lightgray}{34.60} & \textcolor{lightgray}{9.51} & \textcolor{lightgray}{38.50} & \textcolor{lightgray}{22.08} & \textcolor{lightgray}{39.82} & \textcolor{lightgray}{47.04} & \textcolor{lightgray}{20.45} & \textcolor{lightgray}{22.48} & \textcolor{lightgray}{23.78} & \textcolor{lightgray}{23.00} & \textcolor{lightgray}{27.29} & \textcolor{lightgray}{34.27} & \textcolor{lightgray}{78.32} & \textcolor{lightgray}{36.99} & \textcolor{lightgray}{46.27} & \textcolor{lightgray}{49.71} & \textcolor{lightgray}{35.93} & \textcolor{lightgray}{32.06}   \\ %

                &OccFormer~\cite{zhang2023occformer} &  \textcolor{lightgray}{37.04} & \textcolor{lightgray}{9.15} & 45.84 & \textcolor{lightgray}{18.20} & \textcolor{lightgray}{42.80} & 50.27 & 24.00 & \textcolor{lightgray}{20.80} & \textcolor{lightgray}{22.86} & \textcolor{lightgray}{20.98} & 31.94 & 38.13 & \textcolor{lightgray}{80.13} & \textcolor{lightgray}{38.24} & 50.83 & \textcolor{lightgray}{54.3} & \textbf{46.41} & \textbf{40.15}   \\ %

                & BEVFormer$^{\textcolor{red}{\bullet}}$~\cite{li2022bevformer} & \textcolor{lightgray}{34.97} & \textcolor{lightgray}{7.53} & \textcolor{lightgray}{41.77} & \textcolor{lightgray}{16.39} & 44.06 & \textcolor{lightgray}{48.48} & \textcolor{lightgray}{17.27} & \textcolor{lightgray}{20.01} & 
                \textcolor{lightgray}{23.36} & \textcolor{lightgray}{21.16} & 
                \textcolor{lightgray}{28.88} & \textcolor{lightgray}{35.59} & \textcolor{lightgray}{80.12} & \textcolor{lightgray}{35.35} & 
                \textcolor{lightgray}{47.65} & \textcolor{lightgray}{51.89} & 
                \textcolor{lightgray}{40.68} & \textcolor{lightgray}{34.28} \\ %

                & PanoOcc$^{\textcolor{red}{\bullet}}$~\cite{wang2023panoocc} & 38.11 & 9.75 & \textcolor{lightgray}{45.31} & 22.45 & \textcolor{lightgray}{43.13} & \textcolor{lightgray}{50.19} & \textcolor{lightgray}{22.25} & 27.35 & 
                \textcolor{lightgray}{24.49} & \textbf{25.17} & 
                \textcolor{lightgray}{31.74} & \textcolor{lightgray}{37.95} & 81.74 & 42.29 & 
                \textcolor{lightgray}{50.82} & 54.80 & 
                \textcolor{lightgray}{40.81} & \textcolor{lightgray}{37.14} \\ %

                % \cdashline{2-20}
                % \hhline{~-------------------}
                % \hhline{~-} 
                
                & \cellcolor{gray!20} $\text{\textbf{ProtoOcc}}$~\textbf{(Ours)} &  \cellcolor{gray!20}\textbf{39.01} & \cellcolor{gray!20}\textbf{9.75} & \cellcolor{gray!20}\textbf{46.08} & \cellcolor{gray!20}\textbf{24.34} & \cellcolor{gray!20}\textbf{46.09} & \cellcolor{gray!20}\textbf{52.45} & \cellcolor{gray!20}\textbf{24.21} & \cellcolor{gray!20}\textbf{28.11}& \cellcolor{gray!20}\textbf{24.72} & \cellcolor{gray!20}\textcolor{lightgray}{19.79} & \cellcolor{gray!20}\textbf{32.90} & \cellcolor{gray!20}\textbf{40.50} & \cellcolor{gray!20}\textbf{82.29}& \cellcolor{gray!20}\textbf{43.02} & \cellcolor{gray!20}\textbf{52.47} & \cellcolor{gray!20}\textbf{55.94 } & \cellcolor{gray!20}42.46 & \cellcolor{gray!20}38.13  \\ %

                \midrule

                \multirow{3}{*}{\shortstack{\texttt{Small}}}& BEVFormer$^{\color{ForestGreen}{\bullet}}$~\cite{li2022bevformer} & \textcolor{lightgray}{33.98} & \textcolor{lightgray}{6.75} & 41.67 & \textcolor{lightgray}{13.91} & 41.97 & 48.49 & \textcolor{lightgray}{17.83} & \textcolor{lightgray}{18.01} & 22.19 & \textcolor{lightgray}{19.08} & \textcolor{lightgray}{29.64} & \textcolor{lightgray}{33.23} & \textcolor{lightgray}{79.42} & \textcolor{lightgray}{36.48} & \textcolor{lightgray}{46.82} & \textcolor{lightgray}{49.26} & 39.04 & \textcolor{lightgray}{33.91} \\ %

                & PanoOcc$^{\color{ForestGreen}{\bullet}}$~\cite{wang2023panoocc} & 35.78 & 8.18 & \textcolor{lightgray}{41.60} & 20.79 & \textcolor{lightgray}{41.25} & \textcolor{lightgray}{47.78} & 21.87 & 23.42 & \textcolor{lightgray}{21.03} & 19.29 & 29.71 & 36.10 & 81.20 & 40.00 & 49.22 & 53.94 & \textcolor{lightgray}{38.09} & 34.83  \\ %

                & \cellcolor{gray!20}\textbf{ProtoOcc~(Ours)$^{\textcolor{red}{\bullet}}$} &  \cellcolor{gray!20}\textbf{37.80} & \cellcolor{gray!20}\textbf{9.28} & \cellcolor{gray!20}\textbf{43.64} & \cellcolor{gray!20}\textbf{22.30} & \cellcolor{gray!20}\textbf{44.72} & \cellcolor{gray!20}\textbf{50.07} & \cellcolor{gray!20}\textbf{23.68} & \cellcolor{gray!20}\textbf{25.23} & \cellcolor{gray!20}\textbf{22.77} & \cellcolor{gray!20}\textbf{19.66} & \cellcolor{gray!20}\textbf{30.43} & \cellcolor{gray!20}\textbf{38.73} & \cellcolor{gray!20}\textbf{82.05} & \cellcolor{gray!20}\textbf{42.61} & \cellcolor{gray!20}\textbf{51.68} & \cellcolor{gray!20}\textbf{55.84} & \cellcolor{gray!20}\textbf{41.91} &\cellcolor{gray!20}\textbf{38.05} \\ %

                % \cmidrule(r){2-20}

                \midrule
     
			\multirow{3}{*}{\shortstack{\texttt{Tiny}}}
   
                & BEVFormer~\cite{li2022bevformer} & \textcolor{lightgray}{32.02} & \textcolor{lightgray}{4.86} & \textcolor{lightgray}{39.79} & \textcolor{lightgray}{7.17}  & \textcolor{lightgray}{42.46} & \textcolor{lightgray}{47.10} & \textcolor{lightgray}{18.46}  & \textcolor{lightgray}{13.18}  & \textcolor{lightgray}{17.76} & \textcolor{lightgray}{12.46} & \textcolor{lightgray}{28.74} & \textcolor{lightgray}{33.19} & \textcolor{lightgray}{78.64} & \textcolor{lightgray}{35.36} & \textcolor{lightgray}{45.27} & \textcolor{lightgray}{47.29} & \textcolor{lightgray}{38.93} & \textcolor{lightgray}{33.61}  \\ %
                
                & PanoOcc~\cite{wang2023panoocc} & 33.99 & 6.97 & 39.60 & 18.80 & 40.67 & 45.63 & 18.19 & 21.43 & 19.10 & 16.53 & 25.99 & 35.15 & 80.60 & 38.44 & 49.02 & 52.11 & 36.81 & 32.87 \\ %
                
                & \cellcolor{gray!20}\textbf{ProtoOcc~(Ours)$^{\color{ForestGreen}{\bullet}}$} &\cellcolor{gray!20}\textbf{35.68}  & \cellcolor{gray!20}\textbf{8.33} &  \cellcolor{gray!20}\textbf{40.55} &  \cellcolor{gray!20}\textbf{19.84} & \cellcolor{gray!20}\textbf{42.95} & \cellcolor{gray!20}\textbf{48.08} & \cellcolor{gray!20}\textbf{20.31} & \cellcolor{gray!20}\textbf{22.78}& \cellcolor{gray!20}\textbf{21.21} & \cellcolor{gray!20}\textbf{17.00} & \cellcolor{gray!20}\textbf{28.22} & \cellcolor{gray!20}\textbf{36.60} & \cellcolor{gray!20}\textbf{81.42} & \cellcolor{gray!20}\textbf{41.32} & \cellcolor{gray!20}\textbf{50.16} & \cellcolor{gray!20}\textbf{53.82} & \cellcolor{gray!20}\textbf{38.63} & \cellcolor{gray!20}\textbf{35.35}  \\ %

			\bottomrule
		\end{tabular}
		\vspace{-1.0em}
		\label{tab:nuscocc}
	\end{table*}

\section{Experiments}
\subsection{Experiment Setup}
\myparagraph{Datasets and Evaluation Metrics.}
We train and evaluate our method on two representative tasks in 3D scene understanding: 3D occupancy prediction~(3DOP) and 3D semantic scene completion~(3DSSC). For 3DOP, we utilize the Occ3D-nuScenes~\cite{tian2024occ3d} benchmark. The qualitative result is measured with a mean Intersection-over-Union (mIoU) metric. For 3DSSC, we use the SemanticKITTI~\cite{behley2019semantickitti} benchmark, and measure IoU and mIoU for evaluation. More details of the datasets can be found in Appendix~A.

\myparagraph{Implementation Details.}
As our primary goal is to observe the influences of voxel query resolutions on view transformation, categorize the experimental setting based on query resolutions. For Occ3D-nuScenes, we make three variants: \texttt{Base}, \texttt{Small} and \texttt{Tiny}. \texttt{Base} follows the standard baseline settings: a query size of $200\times 200$ for BEVFormer and $100\times 100\times 16$ for PanoOcc and our ProtoOcc. For \texttt{Small}, we use a query size of $100\times 100$ for BEVFormer, and $50\times 50\times 16$ for PanoOcc and ProtoOcc. Lastly for \texttt{Tiny}, we further reduce the resolution into $50\times 50$ for BEVFormer, and $50\times 50\times 4$ for PanoOcc and ProtoOcc. All three categories are trained with $432\times 800$ sized images for 12 epochs. For SemanticKITTI, we make two variants: \texttt{Base} and \texttt{Small}. Here, we observe how our proposed methods can be applied and improve existing baselines: VoxFormer~\cite{li2023voxformer} and Symphonies~\cite{jiang2024symphonize}. A query size of $128\times 128\times 16$ is used for \texttt{Base}, and $64\times 64\times 8$ is used for \texttt{Small} for all baselines. More implementation details are reported in Appendix~B.1.

\begin{table}[t]
    \centering
    \caption{\textbf{Quantitative results on the SemanticKITTI~\cite{behley2019semantickitti} validation set.} Results of each semantic class are provided in Appendix~D. Performance increases through our method are denoted in \textcolor{blue}{blue}.}
    \vspace{-0.5em}
    \scalebox{0.86}{
    \begin{tabular}{lcccc}
            \toprule
            Model & Pub. & IoU & mIoU\\
            \midrule
            MonoScene~\cite{cao2022monoscene} & CVPR 22 & 36.86 & 11.08 \\
            TPVFormer~\cite{huang2023tri} & CVPR 23 & 35.61 & 11.36 \\
            OccFormer~\cite{zhang2023occformer} & ICCV 23 & 36.50 & 13.46 \\
            HASSC~\cite{wang2024not} & CVPR 24 & 44.82 & 13.48 \\\midrule
            VoxFormer-\texttt{S}~\cite{li2023voxformer} & CVPR 23 & 43.10 & 11.51 \\
            + \cellcolor{gray!20}ProtoOcc & \cellcolor{gray!20}- & \cellcolor{gray!20}43.55~\textcolor{blue}{(+0.35)} & \cellcolor{gray!20}12.39~\textcolor{blue}{(+0.88)} \\
            VoxFormer-\texttt{B}~\cite{li2023voxformer} & CVPR 23 & 44.02 & 12.35 \\
            + \cellcolor{gray!20}ProtoOcc & \cellcolor{gray!20}- & \cellcolor{gray!20}\textbf{44.90}~\textcolor{blue}{(+0.85)} & \cellcolor{gray!20}13.57~\textcolor{blue}{(+1.22)} \\\midrule
            Symphonies-\texttt{S}~\cite{jiang2024symphonize} & CVPR 24 & 41.67 & 13.64 \\
            + \cellcolor{gray!20}ProtoOcc & \cellcolor{gray!20}- & \cellcolor{gray!20}43.02~\textcolor{blue}{(+1.35)} & \cellcolor{gray!20}14.50~\textcolor{blue}{(+0.86)} \\
            Symphonies-\texttt{B}~\cite{jiang2024symphonize} & CVPR 24 & 41.85 & 14.38 \\
            + \cellcolor{gray!20}ProtoOcc & \cellcolor{gray!20}- & \cellcolor{gray!20}42.12~\textcolor{blue}{(+0.27)} & \cellcolor{gray!20}\textbf{14.83}~\textcolor{blue}{(+0.45)} \\
            \bottomrule
    \end{tabular}}
    \vspace{-1.4em}
    \label{tab:semantickitti_quan}
\end{table}

\subsection{Quantitative Analysis}

\myparagraph{3D Occupancy Prediction.} 
Table \ref{tab:nuscocc} presents a quantitative comparison of 3D occupancy prediction between baselines and ProtoOcc. Here, we observe two key insights. First, ProtoOcc achieves the highest mIoU in all query settings. This strongly demonstrates the effectiveness of voxel representation learning through our proposed methods. Specifically, ProtoOcc excels in predicting important road agents~(e.g. pedestrian, car, and bus). Furthermore, by comparing the results within the same color marks~($\textcolor{red}{\bullet}$ and $\textcolor{ForestGreen}{\bullet}$), we notice that ProtoOcc achieves competitive results against the baselines with higher-resolution queries. For example, with 75\% fewer parameters in view transformation, ProtoOcc achieves performance comparable to PanoOcc~\cite{wang2023panoocc} in the larger resolution category. The gaps in mIoU are marginal~(0.31 in \texttt{Small}-\texttt{Base} and 0.1 in \texttt{Tiny}-\texttt{Small}), and our ProtoOcc even outperforms predictions in a number of semantic classes.

\myparagraph{3D Semantic Scene Completion.}
Table \ref{tab:semantickitti_quan} reports the results of 3D semantic scene completion with SemanticKITTI benchmark. As clearly seen from the table, our method can also bring benefits in scene understanding as a plug-and-play. Applying our method enhances the performance of both baselines in all metrics. It is important to note that, except for IoU in VoxFormer, all the smaller-resolution models surpass its larger variant when our ProtoOcc is combined. This implies that ProtoOcc overcomes 87.5\% deficiency of spatial storage through its prototype-aware view transformation and multi-perspective occupancy decoding.

\myparagraph{View Transformation Efficiency.}
Table~\ref{tab:cost} compares the cost of inference time, FLOPs, and the number of parameters during the view transformation stage. Initially, by comparing \text{PanoOcc}-\texttt{Base} and PanoOcc-\texttt{Small}, we observe huge savings from a reduced query resolution. Next, we compare two view transformation methods introduced from DFA3D~\cite{li2023dfa3d} and ProtoOcc on the \texttt{Small} query setting. Although additional processing brings inevitable increases in computations, our method shows significant computational efficiency compared to DFA3D, as it produces better results with less costs in inference time and number of parameters. Above all, our ProtoOcc-\texttt{Small} produces comparable mIoU with \text{PanoOcc}-\texttt{Base} while requiring $60.53\%$, $71.15\%$, and $65.16\%$ less computational costs in each metric, respectively.

\begin{figure*}[t]
\begin{center}
\includegraphics[width=\linewidth]{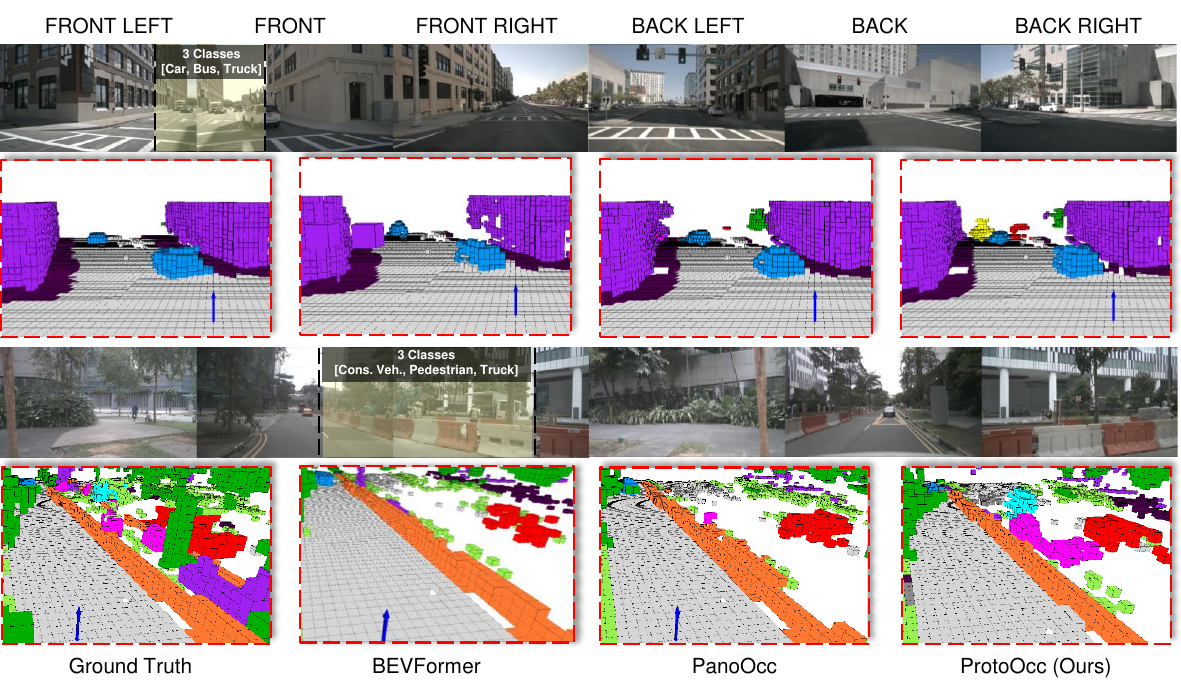}
\end{center}
\vspace{-1.5em}
\caption{\textbf{Predictions in challenging scenarios.} We visualize the prediction comparisons between the baseline and our ProtoOcc. The corresponding camera view is highlighted in yellow, and we label important semantic classes on top. Best viewed in color.}

\label{fig:fig_qual}
\vspace{-1em}
\end{figure*}

\subsection{Qualitative Analysis}
\myparagraph{Predictions on Challenging Scenarios.} 
In this section, we visualize the prediction results on some challenging scenarios that the numerical results were unable to highlight. In the top example of Figure~\ref{fig:fig_qual}, the labels for approaching vehicles in the distance are absent in the ground truth. Accordingly, the baselines fail in predicting the occupancy of the corresponding region. However, our high-level 2D prototype-aware encoding enables ProtoOcc to capture these long-range objects, even without ground-truth guidance. Furthermore, the bottom example emphasizes the superiority of ProtoOcc in occluded regions. We consider that this ability is developed through the multi-perspective learning of 3D space. More examples are provided in Appendix~D.2.

\begin{table}[t]
    \caption{Comparison of Computational Efficiency.}
    \vspace{-0.5em}
    \centering
    \scalebox{0.8}{
    \begin{tabular}{lcccc}
            \toprule
            Model & Inf. Time & FLOPs~(G) & Params & mIoU\\
            \midrule
            \text{PanoOcc}-\texttt{B}~\cite{wang2023panoocc} & 266~ms & 1,310 & 46.24M & 38.11\\
            \midrule
            PanoOcc-\texttt{S} & 66~ms & 330 & 15.52M & 35.78\\
            + DFA3D~\cite{li2023dfa3d} & 153~ms & 360 & 22.33M & 36.27\\\midrule
            \textbf{ProtoOcc--\texttt{S}~(Ours)} & 105~ms & 378 & 16.11M & 37.80\\
            \bottomrule
    \end{tabular}}
    \vspace{-1.3em}
    \label{tab:cost}
\end{table}

\begin{figure}[t]
\begin{center}
   \includegraphics[width=\linewidth]{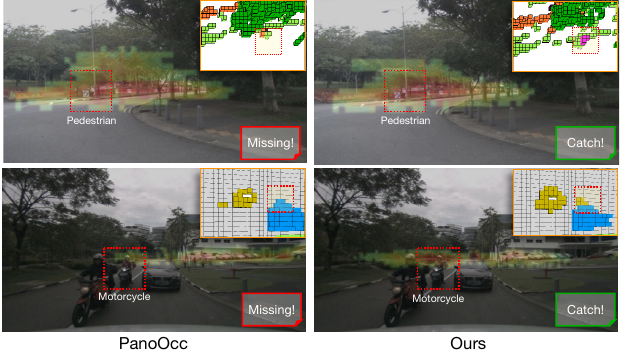}
\end{center}
\vspace{-1em}
\caption{\textbf{Attention map visualization.} Compared to the baseline, ProtoOcc can attend to more important details in the image~(e.g. red dashed boxes), which is crucial for safe driving systems.}
\label{fig:fig_attn}
\vspace{-1em}
\end{figure}

\myparagraph{Attention Map Visualization.} Attention map is a crucial indicator of whether the 2D-to-3D view transformation has been successful. Figure~\ref{fig:fig_attn} compares how different qualities of attention maps can lead to substantial differences in predicted 3D occupancies. The high-level 2D prototypes guide ProtoOcc to distinguish the scene components with clear boundaries. This naturally enables the model to identify important parts in the image to focus on with fine granularity. For example, ProtoOcc captures small yet crucial scene components~(e.g. pedestrian and motorcycle in Figure~\ref{fig:fig_attn}) and assigns high attention weights, which leads to accurate predictions. However, the baseline fails to capture the details and result in prediction misses that jeopardize the safety of the driving environment.

\begin{figure}[t]
\begin{center}
\includegraphics[width=\linewidth]{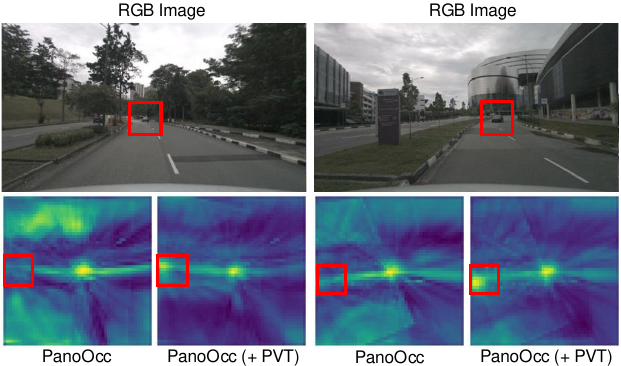}
\end{center}
\vspace{-1em}
\caption{\textbf{Feature map visualization.} We visualize the learned voxel queries, which are average-pooled on the channel and the z-axis. The red box indicates the region we focus on.}
\label{fig:fig_feat_cqe}
\vspace{-1em}
\end{figure}

\myparagraph{Feature Map Visualization.} The prototype-aware view transformation forms a cluster per each significant region in an image. Owing to this strategy, the encoding stage gains flexibility, by not being overwhelmed by visually dominant regions. 
Figure~\ref{fig:fig_feat_cqe} visualizes the feature map of voxel queries, which are flattened in the channel and the z-axis. As clearly seen in the illustration, the features of the baseline are biased towards visually dominant elements in the scene, such as the yard or building. In contrast, applying our prototype-aware view transformation on the baseline effectively enhances the recognition of smaller, yet crucial components for safe driving, such as vehicles.

\begin{table}[t]
    \centering 
    \caption{Ablation on Model Design.}
    \vspace{-0.5em}
     \resizebox{1.0\linewidth}{!}{\begin{tabular}{ccccc}
        \toprule
        & \multicolumn{3}{c}{Method} \\
        \cmidrule(r){2-4}
        Exp. \# & Proto. Mapping & Proto. Optimization & MOD & mIoU\\
        \midrule
        1 & - & - & - & 35.78   \\
        2 & \checkmark & - & - & 35.80   \\
        3 & \checkmark & \checkmark & - & 36.55   \\
        4 & - & - & \checkmark & 37.25   \\
        5 & \checkmark & \checkmark & \checkmark & 37.80   \\
        \bottomrule
    \end{tabular}}
    \vspace{-1.4em}
    \label{tab:module}
\end{table}

\subsection{Ablation Study}

\myparagraph{Model Design.}
Table \ref{tab:module} shows the effects of each method proposed in our ProtoOcc. Exp. 1 is a baseline model without any contribution of our method. When our Prototype Mapping technique is applied~(Exp. 2), we observe a little increase in the performance. However, coupling with Prototype Optimization brings considerable improvements~(Exp. 3). This implies that the explicit objective to refine the 2D prototype quality is highly beneficial for 3D voxel clustering, and as a result helps to predict accurate semantic occupancy. In Exp. 4, we apply the Multi-perspective Occupancy Decoding~(MOD) method on the baseline, which shows great enhancements just by itself. This demonstrates that the contextual diversity is crucial for predicting higher-dimension 3D occupancy with lower-dimension 2D images. Lastly, the full ProtoOcc model~(Exp. 5) leverages the synergies of each contribution, resulting in superior performance.

\begin{comment}

\end{comment}

\begin{table*}[ht!]
    \scriptsize
    \setlength{\tabcolsep}{0.018\linewidth} 
    \caption{Effects of different augmentations and their consistency regularization~(C.R.).}
    \centering
    % \scalebox{0.75}{
    \begin{tabular}{cc|ccccccccccc}
    \toprule
    &Exp. \# & 1 & 2 & 3 & 4 & 5 & 6 & 7 & 8 & 9 & 10 & 11 \\ \midrule 
      \multirow{4}{*}{Augmentation}  & Random Dropout&- & \checkmark&- &- &- &\checkmark &\checkmark &- &- &\checkmark &-\\
         &  Gaussian Noise &-&- & \checkmark&- &- & -&- &\checkmark&\checkmark&\checkmark &-\\ \cmidrule(r){2-13}
         &  Transpose&- &- &- &\checkmark &- &\checkmark &- &\checkmark &-&- &\checkmark\\ 
         &  Flips&- &- &- &- &\checkmark &- &\checkmark &- &\checkmark&- &\checkmark\\ \midrule 
         \multirow{2}{*}{mIoU} & w/o C.R. &35.78 & 36.99&36.46 &36.49 &36.70 &36.95 &36.78 &36.47 &36.72 & 36.66&36.59\\
         & w/ C.R. &N/A &\textbf{37.19} &\textbf{36.80} &\textbf{36.86} &\textbf{37.18} & \textbf{37.21}& \textbf{37.16}& \textbf{36.90}& \textbf{36.82}&\textbf{37.25} &\textbf{36.78}\\
    \bottomrule
    \end{tabular}
    % \vspace{-0.5em}
    \label{tab:aug}
\end{table*}

\begin{table}[!]
    \centering
    \caption{Sensitivity on the number of 2D prototypes $M$.}
    \vspace{-0.5em}
    \resizebox{1.0\linewidth}{!}{
    \begin{tabular}{lcccc}
        \toprule
         2D prototype $M$ & 1440 $(r=2)$ & 350 $(r=4)$ & 144 $(r=6)$ & 91 $(r=8)$ \\
         \cmidrule{1-5}
         \# Pseudo Mask & \multicolumn{4}{c}{299} \\
         mIoU & 36.90 & 37.80 & 36.95 & 37.07  \\
        \bottomrule
     \end{tabular}}
     % \vspace{-0.3em}
    \label{tab:abl_cluster}
\end{table}

% \multicolumn{4}{c}{299} 
\begin{table}[!]
    \centering
    \caption{Effect of pseudo mask quality.}
    \vspace{-0.5em}
    \resizebox{1.0\linewidth}{!}{
    \begin{tabular}{cccc}
        \toprule
         2D prototype $M$ & Mask Generator & \# Pseudo Mask & mIoU \\
        \midrule
         \multirow{2}{*}{350} & SEEDS~\cite{van2012seeds} & 91 & 37.14  \\
         & Segment Anything~\cite{kirillov2023segment} & 92 & 37.29 \\
        \bottomrule
     \end{tabular}}
     % \vspace{-0.3em}
    \label{tab:abl_quality}
\end{table}

\begin{table}[!]
    \centering
    \caption{Impact of the granularity level.}
    \vspace{-0.5em}
    \resizebox{1.0\linewidth}{!}{
    \begin{tabular}{cccc}
        \toprule
         2D prototype $M$ & Mask Generator & \# Pseudo Mask & mIoU \\
         \midrule
         \multirow{2}{*}{91} & SEEDS~\cite{van2012seeds} & 299 & 37.07  \\
         & Segment Anything~\cite{kirillov2023segment} & 92 & 37.21 \\
         \multirow{2}{*}{350} & SEEDS~\cite{van2012seeds} & 299 & 37.80  \\
         & Segment Anything~\cite{kirillov2023segment} & 92 & 37.29 \\
        \bottomrule
     \end{tabular}}
     % \vspace{-0.3em}
    \label{tab:abl_granularity}
\end{table}

\myparagraph{Prototype Analysis.}
In this section, we deeply analyze the Prototype-aware View Transformation through three experiments: 1) sensitivity on number of 2D prototypes $M$, 2) choice for pseudo mask generator, and 3) impact of granularity level.
In the first experiment, we fix the pseudo mask generator as SEEDS~\cite{van2012seeds} and observe the performance sensitivity on $M$, which is determined by the pre-defined downsampling ratio $r$. As shown in Table~\ref{tab:abl_cluster}, we discover that overly fine ($r$=2) or coarse ($r$=8) clustering leads to suboptimal performance and that $M$=350 with $r$=4 is the adequate number of prototypes. Second, Table~\ref{tab:abl_quality} shows that the pseudo mask generator SAM~\cite{kirillov2023segment}, which generates more semantically consistent clusters, outperforms the traditional superpixel algorithm SEEDS under consistent number of pseudo masks.
Lastly, however, we observe in Table~\ref{tab:abl_granularity} that aligning the consistent number of 2D prototypes $M$ with the number of pseudo mask is more crucial for facilitating the Prototype Optimization.

\myparagraph{Augmentations and Consistency Regularization.}
ProtoOcc introduces both feature- and spatial-level voxel augmentation methods for diversifying 3D contexts. Table \ref{tab:aug} shows the influences of each augmentation and their combinations, with and without scene consistency regularization. Exp. 1 is the baseline model just as in Table \ref{tab:module}. The results of the experiment highlight three key insights. First, the augmentation itself brings a considerable amount of benefits, demonstrating the importance of contextual diversity for occupancy decoding. Secondly, further regularization on their output consistency guarantees they represent the same scene, thereby naturally boosting robustness. Lastly, incorporating numerous augmentations for diversity can occasionally hinder robust feature learning and doesn't consistently lead to proportional performance improvements. For this reason, we conduct empirical experiments with no more than two combinations of augmentations.

\section{Conclusion}
This paper introduces ProtoOcc, an innovative method addressing the challenge of enriching contextual representation in low-resolution voxel queries for camera-based 3DOP. ProtoOcc enhances the interpretability of smaller-sized queries while mitigating information loss through two context-enriching strategies: 1) a prototype-aware view transformation and 2) a multi-perspective occupancy decoding. Experimental evaluations on both Occ3D and SemanticKITTI benchmark underscore the significance of ProtoOcc, revealing substantial enhancements over previous approaches. Notably, ProtoOcc achieves competitive performance even with smaller-sized queries, offering a promising solution for real-time deployment without compromising predictive accuracy.

\section*{Acknowledgement}
{
This work was primarily supported by Samsung Advanced Institute of Technology (SAIT) (50\%),
the Culture, Sports and Tourism R\&D Program through the Korea Creative Content Agency grant funded by the Ministry of Culture, Sports and Tourism in 2024~(Research on neural watermark technology for copyright protection of generative AI 3D content, RS-2024-00348469, 25\%; International Collaborative Research and Global Talent Development for the Development of Copyright Management and Protection Technologies for Generative AI, RS-2024-00345025, 14\%),
the National Research Foundation of Korea~(NRF) grant funded by the Korea government~(MSIT)(RS-2025-00521602, 10\%), and
Institute of Information \& communications Technology Planning \& Evaluation~(IITP) grant funded by the Korea government~(MSIT) (No. RS-2019-II190079, Artificial Intelligence Graduate School Program~(Korea University), 1\%), and 
Artificial intelligence industrial convergence cluster development project funded by the Ministry of Science and ICT~(MSIT, Korea)\&Gwangju Metropolitan City.
}
{
    \small
    \bibliographystyle{ieeenat_fullname}
    \bibliography{main}
}

% \onecolumn
% \appendix 
% \input{sec/X_suppl}
\end{document}